\documentclass[manuscript,screen,a4paper]{acmart}

\usepackage[utf8]{inputenc}
\usepackage{subcaption}
\usepackage{pgfplots}
\usepackage{booktabs}
\usepackage{makecell}
\usepackage{multirow}
\usepackage{longtable}
\usepackage{siunitx}
\usepackage{caption}
\usepackage{lscape}
\usetikzlibrary{patterns}
\usepackage{array}
\usepackage{tabularx}
\usepackage{float}

\AtBeginDocument{%
  \providecommand\BibTeX{{%
    \normalfont B\kern-0.5em{\scshape i\kern-0.25em b}\kern-0.8em\TeX}}}

\setcopyright{acmlicensed}
\copyrightyear{2024}
\acmYear{2024}
\acmDOI{XXXXXXX.XXXXXXX}

\acmConference[Conference acronym 'XX]{Make sure to enter the correct
  conference title from your rights confirmation emai}{June 03--05,
  2018}{Woodstock, NY}
\acmISBN{978-1-4503-XXXX-X/18/06}

\begin{document}

\title{Introducing WARM-VR: A Benchmark Dataset for Multimodal Wearable Affect Recognition in Virtual Reality}

\author{Karim Alghoul}
\email{Karim.Alghoul@uottawa.ca}
\orcid{0009-0007-6792-1399}
\affiliation{%
  \institution{School of Electrical Engineering and Computer Science, University of Ottawa}
  \city{Ottawa}
  \state{Ontario}
  \country{Canada}
}

\author{Mohd Faisal}
\email{mmohd055@uottawa.ca}
\orcid{0000-0003-3363-4758}
\affiliation{%
  \institution{School of Electrical Engineering and Computer Science, University of Ottawa}
  \city{Ottawa}
  \state{Ontario}
  \country{Canada}
}

\author{Fedwa Laamarti}
\email{flaamart@uottawa.ca}
\orcid{0000-0002-0338-9264}
\affiliation{%
  \institution{Computer Vision Department, Mohamed bin Zayed University of Artificial Intelligence}
  \city{Abu Dhabi}
  \country{UAE}
}
\affiliation{%
  \institution{School of Electrical Engineering and Computer Science, University of Ottawa}
  \city{Ottawa}
  \state{Ontario}
  \country{Canada}
}

\author{Hussein Al Osman}
\email{hussein.alosman@uottawa.ca}
\orcid{0000-0002-7189-5644}
\affiliation{%
 \institution{School of Electrical Engineering and Computer Science, University of Ottawa}
  \city{Ottawa}
  \state{Ontario}
  \country{Canada}
}

\author{Abdulmotaleb El Saddik}
\email{elsaddik@uottawa.ca}
\orcid{0000-0002-7690-8547}
\affiliation{%
 \institution{School of Electrical Engineering and Computer Science, University of Ottawa}
  \city{Ottawa}
  \state{Ontario}
  \country{Canada}
}

\renewcommand{\shortauthors}{K.Alghoul et al.}

\begin{abstract}
  With the growing integration of human-computer interaction into everyday life, advances in machine learning have enabled systems to better perceive and respond to users’ emotional states. Affective computing, in particular, has emerged as a key area of research, leveraging continuous physiological signals from wearable sensors to detect and measure human affect. However, most existing affect recognition datasets focus on static environments, limiting their applicability to immersive multimedia contexts such as Virtual Reality (VR). In this paper, we introduce WARM-VR (Wearable Affect Recognition from Multisensory stimuli in Virtual Reality), a novel publicly available multimodal dataset designed to support affect recognition in immersive, multisensory environments using wearable sensing instrumentation. Data were collected from 31 participants aged 19–37 using wearable sensors: a wristband measuring Blood Volume Pulse (BVP), Electrodermal Activity (EDA), skin Temperature (TEMP), three-axis Acceleration (ACC), and a chest strap recording Electrocardiogram (ECG) signals. Participants engaged in immersive VR experiences designed to elicit relaxation through a calming beach environment following stress induction via an arithmetic task. These sessions incorporated synchronized multimedia stimuli: visual, auditory, and olfactory. Affective states were assessed subjectively through validated self-report questionnaires and objectively through the analysis of physiological measurements. Statistical analysis of the questionnaires confirmed that VR relaxation significantly reduced negative affect, particularly with olfactory enhancement. Furthermore, we established a benchmark on the dataset using widely recognized machine learning algorithms. The best performance for binary classification from BVP data of valence, was obtained with a CNN and a CNN–Bi-GRU model, both achieving an average F1-score of 0.63 and an AUC of 0.69. For arousal, a lightweight Transformer architecture provided the most balanced results (F1-0 0.54 and F1-1 0.63), outperforming recurrent hybrids. In the relaxation task, a CNN–Bi-GRU model reached the highest overall performance (average F1-score 0.64, AUC 0.69), with the Transformer variant achieving comparable accuracy. The WARM-VR dataset in this paper is made publicly available and can be downloaded from: DOI \href{https://ieee-dataport.org/documents/warm-vr-wearable-affect-recognition-multisensory-stimuli-virtual-reality-dataset}{10.21227/rwrm-n531} 
\end{abstract}



\received{20 February 2007}
\received[revised]{12 March 2009}
\received[accepted]{5 June 2009}
\maketitle

\section{Introduction}
Human-computer interaction has become integral to daily life, particularly with the rapid advancement of machine learning technologies \cite{Lv2024}. To further improve the quality of these interactions, the field of affective computing has emerged, an interdisciplinary domain focused on designing systems that can recognize, interpret, and respond to human emotions \cite{Zhang2025}. Specifically, Affective Computing in Multimedia (ACM) focuses on recognizing the emotional responses that a particular stimuli is likely to elicit in users \cite{Zhao2019}.

In recent years, the rise of wearable devices has provided a valuable new data source for mental well-being and affect recognition \cite{Yin2022} \cite{Chen2022} \cite{Vahdati2025}. These devices enable continuous, non-intrusive monitoring of physiological signals such as BVP and ECG, which are commonly used to infer affective states \cite{Rashed2025}. Unlike facial expressions or vocal tone, physiological signals offer objective and reliable indicators of affect, as they are less susceptible to social masking \cite{Zhu2019}\cite{Zhao2019v2} \cite{Latifzadeh2024} \cite{Ahmad2022}. These advances have inspired the development of datasets aimed at modeling affective states from physiological signals. However, most are constrained by non-immersive environments. 

VR is an effective tool for simulating real-life environments, enabling users to interact with digital surroundings and experience a strong sense of presence despite not being physically there. It has gained increasing recognition for its potential to promote relaxation \cite{CamaraD} and reduce anxiety \cite{Grenier2015} through the integration of multiple sensory modalities. This integration has been shown to enhance the depth of presence and realism in VR systems \cite{CamaraD}. While visual and auditory stimuli have traditionally dominated VR experiences, olfactory stimuli is increasingly being explored for its potential to modulate mood and improve emotional well-being \cite{Igarashi2014}. The use of scent in virtual settings, similar to aromatherapy, involves the controlled dispersion of essential oils known for their calming properties. For instance, lavender and orange scents have been shown to reduce anxiety and elevate mood \cite{Igarashi2014}.

Despite these advances, existing affect recognition datasets remain limited in several important aspects. Most lack full immersion due to the absence of VR integration, even when multiple sensory stimuli are included. Additionally, physiological data are often collected using stationary lab equipment rather than wearable devices, which undermines real world applications. In contrast, WARM-VR is, to the best of our knowledge, the first dataset specifically designed for affect recognition from wearable devices within a fully immersive virtual environment that also incorporates olfactory stimulation.

The main contributions of this paper are:
\begin{enumerate}
  \item We present a new publicly available, multimodal dataset (vision, sound, and olfaction) conducted in an immersive VR environment. Two wearable devices are used: a wrist-based sensor and a chest-based belt, capturing high-resolution physiological signals (BVP, ECG, EDA, TEMP) and motion signal (ACC). The dataset also includes self-reported data of participants’ affective states, collected through questionnaires. These responses can serve as valuable labels for training personalized affect recognition models.
  \item We provide benchmark results using Heart Rate Variability (HRV) analysis from ECG signals, along with widely used deep learning models, including CNN, LSTM, GRU and Transformers, to support future research in affect recognition.
\end{enumerate}

\section{RELATED WORK}
Affective computing research has long emphasized the value of physiological signals for understanding affective responses to multimedia stimuli \cite{Li2022}. Numerous publicly available datasets have been proposed to support this goal, often incorporating modalities such as Electroencephalogram (EEG), Electromyogram (EMG), ECG, BVP, and EDA. However most existing datasets were collected in controlled lab conditions using non-immersive visual stimuli and do not consider the growing relevance of immersive VR and multisensory environments. Our work builds on this trajectory by introducing a multimodal dataset based on wearable sensing within immersive VR environments, enhanced with olfactory stimulation.

Among popular datasets, DEAP \cite{Koelstra2012} is a publicly available multimodal database designed to support the analysis of human affective states. It was created to facilitate research in affect recognition by providing physiological and multimedia data collected while participants experienced emotional stimuli. The focus of the DEAP dataset was the EEG signals, but it also included other modalities such as BVP. Despite its broad modality coverage and self-reports, the sensors used are not considered wearables. It also lacks immersion and uses screen-based stimuli only. In contrast, WESAD \cite{Schmidt2018} employs wearable sensors through a chest belt and a wristband to record ECG, BVP, EDA, TEMP, EMG, respiration (RESP), and ACC signals to measure affect-related physiological responses. While WESAD is highly relevant to our focus on wearables, it is limited to visual and sound stimuli and does not consider immersive environments. 

Several datasets have explored the integration of olfactory stimuli into affective elicitation. The OVPD dataset \cite{Xue2022} for example, combines video and odor cues to enhance EEG-based affect classification, showing that multimodal sensory input can improve signal discriminability. Similarly, the DEAR-MULSEMEDIA dataset \cite{Raheel2021} includes EEG, GSR, and BVP signals collected under visual, auditory, olfactory, haptic, and thermal stimuli. This dataset is notable for its multisensory scope, yet it lacks a VR-based or fully immersive delivery context. The influence of BVP, also referred to as photoplethysmography (PPG), on affect analysis has also been explored in isolation. For example, the dataset introduced by Jin et al. \cite{Jin2022} focuses exclusively on PPG-based affect classification using narrative videos and provides benchmark results for a deep learning model. Cognitive AR-based experiments have also been used to investigate affective engagement. Dasdemir introduced BOOKAR \cite{Dademir2022}, a dataset using EEG data collected while participants read texts augmented reality (AR). Although this dataset moves toward immersive interaction, it does not include VR, wearables beyond EEG, or multisensory integration.

Immersive VR offers a compelling way to simulate real-world environments and scenarios, allowing users to feel a strong sense of presence and interact with the environment. By replacing the physical environment with a virtual one, VR creates the illusion of “being there” through sensory engagement. This sense of immersion is largely driven by the stimulation of various senses. By engaging multiple senses, VR can closely replicate real-life interactions. Although often associated with gaming, recent studies have highlighted VR’s potential in therapeutic contexts, including its ability to support relaxation and alleviate anxiety \cite{Benbow2019}.

In 2023, Dasdemir introduced VREMO \cite{Dademir2023} dataset, which includes EEG signals in response to VR scenes to classify both cybersickness and emotional states within the valence–arousal space. This work shows the potential benefit of physiological measurements in immersive contexts. However, it relies solely on EEG and lacks common daily wearables like ECG belts and PPG wristbands. 

We summarize and compare key datasets in Table \ref{table:1} to highlight the novelty of the WARM-VR dataset. To the best of our knowledge, WARM-VR is the first publicly available dataset to combine data from everyday wearables with olfactory stimuli in an immersive VR setting tailored for affect detection. It advances prior work by bridging wearable multimodal affect sensing with fully immersive, multisensory environments such as those envisioned in the metaverse.

\begin{table*}
  \caption{COMPARISON OF PUBLIC AFFECT RECOGNITION DATASETS RELEVANT TO WARM-VR}
  \label{table:1}
  \resizebox{\textwidth}{!}{%
  \begin{tabular}{lcccl}
    \toprule
    \textbf{Dataset} & \textbf{VR-Based} & \textbf{Olfactory} & \textbf{Use of Wearables} & \textbf{Multimodal Focus}\\
    \midrule
    DEAP ~\cite{Koelstra2012} & No & No & No & EEG, BVP, GSR, EMG, TEMP \\
    WESAD ~\cite{Schmidt2018} & No & No & \textbf{Yes} & ECG, EMG, RESP, BVP, EDA, ACC, TEMP \\
    OVPD ~\cite{Xue2022} & No & \textbf{Yes} & No & EEG \\
    DEAR-MULSEMEDIA ~\cite{Raheel2021} & No & \textbf{Yes} & No & EEG, BVP, GSR \\
    Jin et al. ~\cite{Jin2022} & No & No & No & BVP \\
    BOOKAR ~\cite{Dademir2022} & No (AR) & No & No & EEG \\
    VREMO ~\cite{Dademir2023} & \textbf{Yes} & No & No & EEG \\
    \textbf{WARM-VR (Ours)} & \textbf{Yes} & \textbf{Yes} & \textbf{Yes} & \textbf{ECG, BVP, EDA, ACC, TEMP} \\
    \bottomrule
  \end{tabular}
  }
\end{table*}

\section{Data Collection}
\begin{figure}[t]
  \centering
  \includegraphics[width=\linewidth]{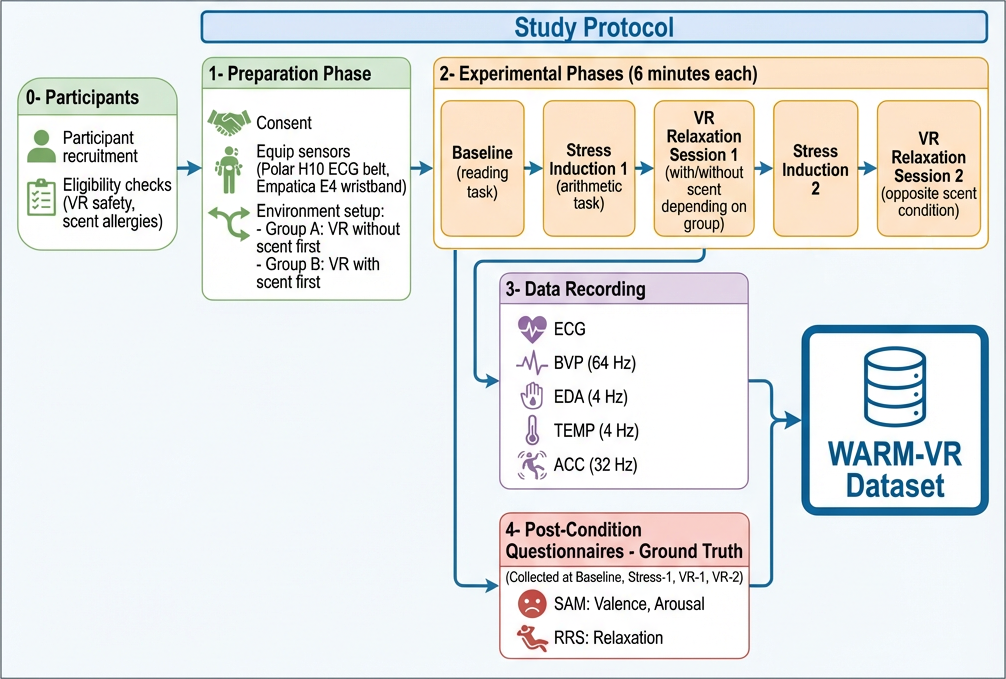}
  \caption{Overview of the Dataset Creation Process}
  \Description{}
  \label{fig:dataset}
\end{figure}

The process of the data collection is shown in Figure \ref{fig:dataset}.

\subsection{Participants}

The study involved 31 participants (13 female, 17 male, 1 preferred not to specify) between the ages of 19 and 37 (Mean Average = 26.6). Three participants were excluded due to an error in sensor placement, which affected the data. A significant focus of our dataset was ensuring diversity among the participants, an aspect that has been largely overlooked in popular existing datasets where the majority of participants were from a single background \cite{Koelstra2012} \cite{Schmidt2018} \cite{Jin2022}. To achieve this, our participants were from various ethnic backgrounds (Figure \ref{fig:1}). This inclusion of participants from multiple ethnic backgrounds enhances the dataset’s demographic diversity, which is important for developing more robust and generalizable affect recognition models. While the sample size is limited and some groups remain underrepresented, it offers broader representation than many existing datasets in the field.
To ensure safety during the VR simulation, individuals with a history of seizures, epilepsy, severe motion sickness, or other neurological conditions were excluded \cite{Birckhead2019}. Participants with known sensitivities or allergies to the scents used in the experiment were also not eligible. 
Recruitment was primarily carried out through posters placed around the University of Ottawa campus. Additional participants were recruited using a snowball sampling method, where initial participants were invited to refer friends and acquaintances to the study.

\subsection{Equipment used}

To ensure accurate and high-resolution physiological data collection in a fully immersive environment, a combination of wearable sensors and sensory stimulation tools was employed. The selected equipment enabled synchronized recording of cardiac activity, delivery of olfactory stimuli, and presentation of a visually and auditorily rich virtual environment. Below is an overview of the key devices used in the study:

\begin{enumerate}
  \item \textbf{ECG Belt:}
  Cardiac electrical activity was measured using the Polar H10 chest strap \cite{polar}, which participants wore around their ribcage. The device uses two electrode pads to capture ECG signals and records R-R intervals with a sampling rate of 1000 Hz. The R-R intervals data were transmitted in real time via Bluetooth.
  \item \textbf{Wristband:}
  Participants also wore the Empatica E4 wristband \cite{e4} on their non-dominant hand. This wearable device continuously recorded multiple physiological signals: BVP at 64 Hz, EDA at 4 Hz, TEMP at 4 Hz, and ACC at 32 Hz. The E4 is widely used in affective computing and clinical research \cite{kutt} due to its high signal fidelity and unobtrusive design \cite{Milstein2020}.
  \item \textbf{VR Headset:}
  To create an immersive virtual environment, we employed the Meta Quest 2 headset \cite{metaQ}. Its high-resolution display and integrated motion tracking enabled participants to engage with the VR scenario naturally while remaining seated. The headset’s wireless design helped eliminate external distractions.
  \item \textbf{Olfactory Diffuser:}
  Olfactory stimulation was delivered via an ultrasonic diffuser that emitted a mist of essential oils chosen to evoke a beach-like scent. The diffuser was activated prior to the participant’s entry, ensuring a consistent ambient aroma throughout the session. This passive exposure enhanced immersion without requiring any participant interaction.
\end{enumerate}

\subsection{Study Protocol}

\begin{figure}[t]
  \centering
  \includegraphics[width=\linewidth]{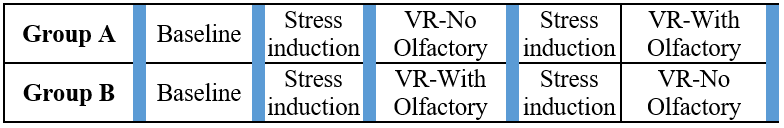}
  \caption{Overview of the two study protocol groups. Each phase lasted 6 minutes. Blue boxes indicate the time points when participants completed questionnaires}
  \Description{}
  \label{fig:3}
\end{figure}

The objective of the study was to induce and measure three different affective states: Baseline, Stress, and Relaxation. As shown in Figure \ref{fig:3}, participants followed a structured sequence of phases in one of two protocol groups (Group A or Group B), depending on the order in which they were exposed to olfactory stimuli. Below are the different phases of our study protocol:

\begin{figure}[t]
  \centering
  \includegraphics[width=0.6\linewidth]{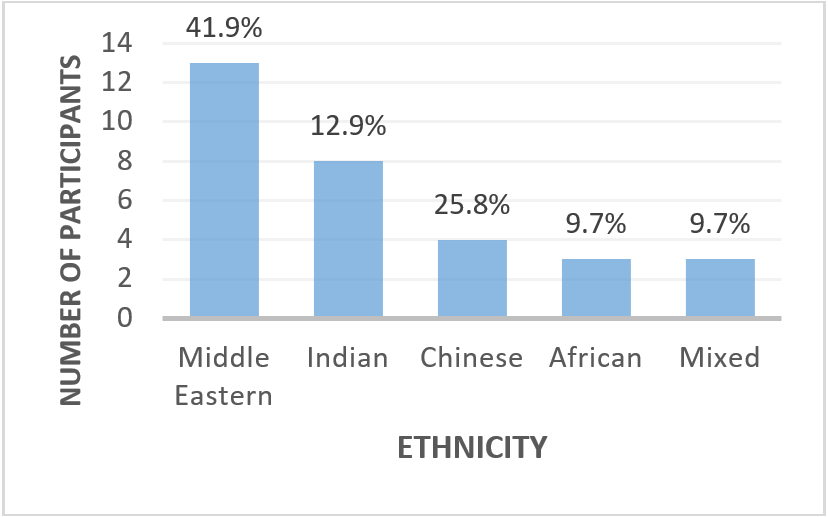}
  \caption{Ethnicity Distribution of Participants}
  \Description{A graph that contains the percentage of each ethnicity participated in the experiment.}
  \label{fig:1}
\end{figure}

\begin{figure}[t]
  \centering
  \includegraphics[width=\linewidth]{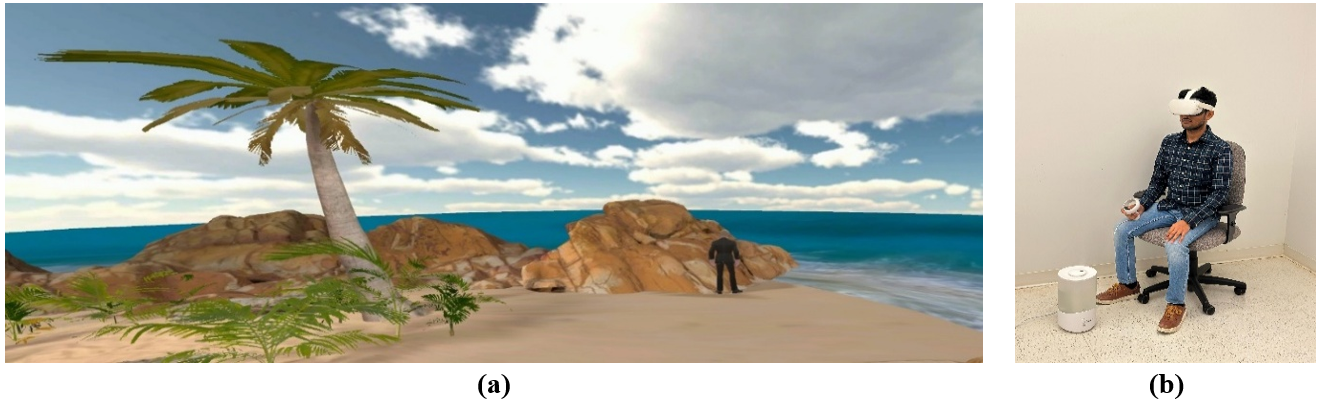}
  \caption{(a) Virtual beach environment with an avatar inside the scene (b) A participant seated in a chair wearing a Meta Quest VR headset, with a diffuser dispersing beach scent to enhance immersion}
  \Description{first image is a screenshot from the metaverse beach environment. The second image is of one of the participants wearing a VR headset and sitting on the chair.}
  \label{fig:2}
\end{figure}

\begin{enumerate}
  \item \textbf{Preparation Phase:}
  The study was conducted in two controlled office spaces: one scent-free for Group A and the other infused with a beach scent for Group B, allowing for controlled exposure to olfactory stimuli. Upon arrival, each participant provided informed consent and completed a pre-experiment questionnaire. They were then fitted with the two wearable devices: a chest-worn ECG belt (Polar H10) and an Empatica E4 wristband, to be used in all the next phases to record the physiological data. Participants were instructed to refrain from alcohol for 12 hours, and from caffeine, food, smoking, and exercise for 3 hours to ensure consistent physiological baselines.
  \item \textbf{Baseline Phase:}
  Participants were seated comfortably and asked to read The Silk Roads by Peter Frankopan for six minutes to establish a resting baseline. Reading was selected over personal device use to minimize external stressors and ensure consistency.
  \item \textbf{Stress Induction Phase:}
  Participants completed a six-minute arithmetic stress task adapted from \cite{AlOsman2016}, involving rapid-fire multiplication problems (numbers from 0 to 12) on a screen with a time limit for each question. Incorrect answers led to score penalties displayed in real time. To increase cognitive pressure, participants were told that their scores were compared to other participants.
  \item \textbf{VR Relaxation Phase – First Session:}
  After the first stress phase, participants were fitted with the Meta Quest VR headset and immersed in a virtual beach environment as seen in Figure \ref{fig:2}. Depending on their group (as shown in Figure \ref{fig:3}), this first session either included olfactory stimuli (Group B) or not (Group A). The VR session lasted six minutes, during which participants were encouraged to explore the virtual scene naturally while remaining seated.
  \item \textbf{Second Stress phase and Second VR session:}
  Participants then completed a second arithmetic stress task identical to the first to re-induce stress, followed by a second VR session in the opposite olfactory condition. Group A, for example, experienced scent-free VR first and scented VR second, while Group B experienced the reverse. This crossover design ensured that the influence of olfactory stimuli could be evaluated independently of the order.
  \item \textbf{Post-Condition Questionnaires:}
  Participants completed the Self-Assessment Manikin (SAM) and Relaxation Rating Scale (RRS) questionnaires after the baseline phase, first stress induction phase, and after each VR relaxation phase (as indicated in \ref{fig:3}). To reduce the study’s time and not burden the participants, questionnaires were not repeated after the second stress induction phase. Answers from the questionnaires were used as subjective labels for participants’ affective states.
\end{enumerate}

\subsection{Ground Truth}
\label{ground-truth}

\begin{figure}[t]
  \centering
  \includegraphics[width=\linewidth]{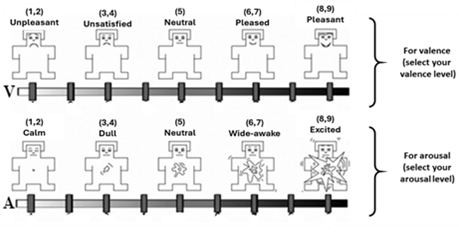}
  \caption{Image shown to participants to help them answer the SAM questionnaire}
  \Description{}
  \label{fig:4}
\end{figure}

To evaluate participants' affective states throughout the experiment, a combination of self-report tools was used to capture both affective responses. These subjective measures served as the ground truth for labeling affective states associated with each experimental phase:

\begin{enumerate}
  \item \textbf{SAM:}
  Affective states were measured using the SAM, a non-verbal pictorial tool that assesses affect along three dimensions: valence (Unpleasant–Pleasant), arousal (Calm–Excited), and dominance (Control). In this study, we excluded dominance as it is not commonly used in literature. In the valence-arousal space, affect can be mapped into four quadrants (e.g., high valence and high arousal may indicate excitement or happiness). Participants used visual aids (see Figure \ref{fig:4}) to rate their affective state on both dimensions.
  \item \textbf{RRS:}
  Participants also self-reported their level of relaxation using the RRS. Participants rated their perceived relaxation on a scale from 0 (not relaxed at all) to 10 (extremely relaxed).
  \item \textbf{VR Relaxation Sessions:}
  To gain insight into participants’ subjective experience with the VR environment, two additional post-experiment questions were included: (1) \textit{“Would you use this method as a relaxation technique?”}  and (2) \textit{“Did you feel more immersed with the scent compared to without the scent?”} These questions aimed to assess both the perceived effectiveness of the olfactory-enhanced VR setup and the depth of user immersion, providing valuable context for future real-world applications.
\end{enumerate}

\section{Use Cases and Methods}

The primary goal of this dataset is to explore how immersive environments, specifically those incorporating VR and olfactory stimuli, can support relaxation and reduce stress. Additionally, the dataset was intentionally designed with broader applicability in mind by collecting a diverse set of physiological signals and self-reported measures.

\subsection{Use Cases}
\begin{enumerate}
  \item \textbf{Investigating the Effect of Olfactory Stimuli:}
  Because each participant experienced both scented and unscented VR conditions in a counterbalanced design, this dataset allows researchers to isolate and analyze the specific impact of olfactory input on relaxation and stress. The inclusion of paired conditions supports within-subject comparisons and enables analysis of the physiological and subjective differences associated with scent-enhanced immersive experiences.
  \item \textbf{HRV Analysis:}
  The ECG data collected using the Polar H10 chest strap include precise R-R interval measurements, which are ideal for traditional HRV analysis. HRV is a well-established method for assessing autonomic nervous system activity, particularly in stress research. Frequency-domain features, such as the high-frequency (HF) component (0.15–0.4 Hz), have been shown to corelate strongly with relaxation states \cite{SAKAKIBARA1994}. Researchers can extract these features to evaluate physiological responses across different experimental conditions.
  \item \textbf{Affect Measurement via Machine Learning:}
  The combination of physiological data from both the ECG belt and the E4 wristband, along with subjective relaxation scores from the RRS questionnaire (0–10 scale), offer a robust foundation for training machine and deep learning models aimed at relaxation estimation. Additionally, valence and arousal ratings collected via the SAM questionnaire (see Section 3.4) allow for broader affective modeling within the dimensional affective space. These multimodal labels support applications in biofeedback, wellness monitoring, and personalized affect-aware systems.
  \item \textbf{Statistical Analysis of Subjective Responses:}
  The dataset includes several validated psychological questionnaires, including SAM, and RRS, administered at multiple time points. These tools can be used for traditional statistical analysis to explore the effects of different experimental conditions on mood. This makes the dataset useful not only for signal processing and machine learning but also for hypothesis-driven research in psychology and human-computer interaction.
\end{enumerate}

\subsection{Methods}
\begin{figure}[t]
  \centering
  \includegraphics[width=\linewidth]{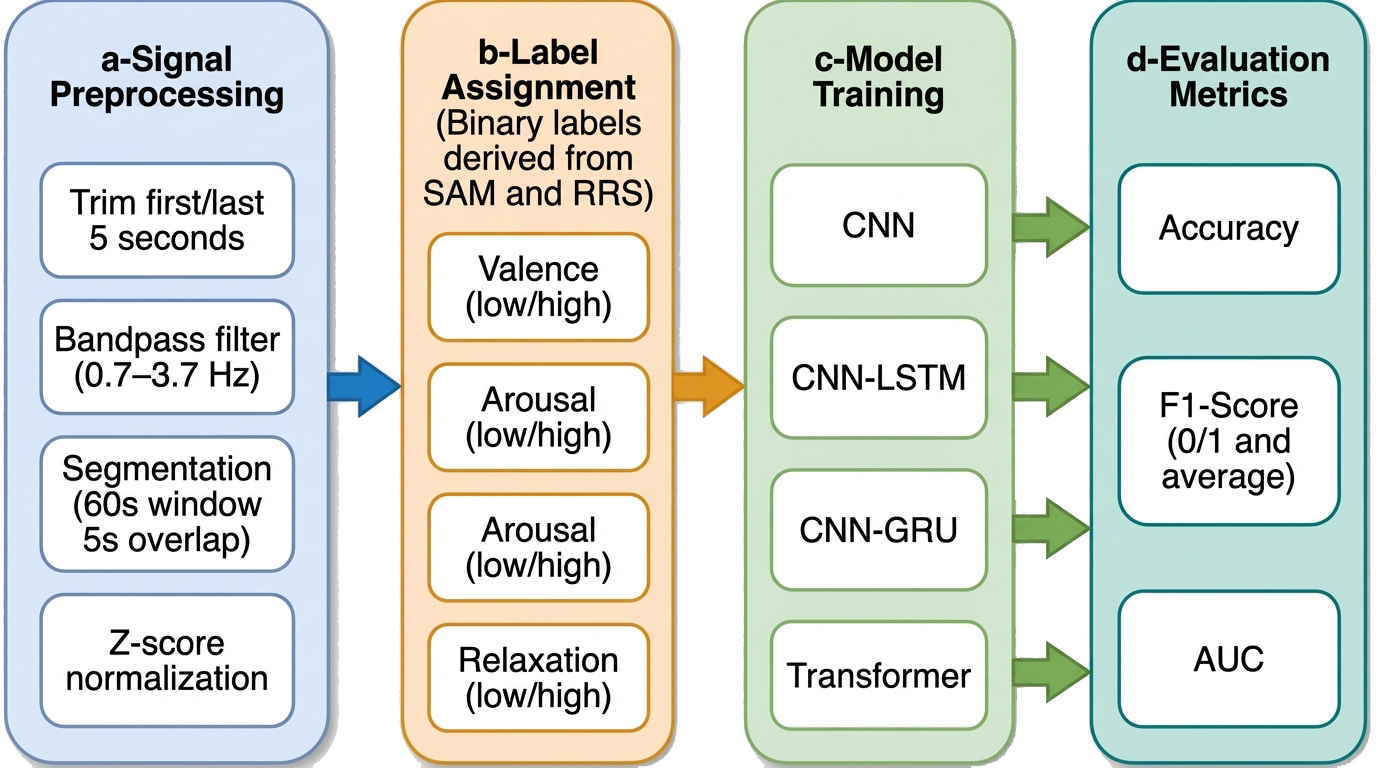}
  \caption{Workflow of the Machine Learning Validation Pipeline for Affect Recognition Using the WARM-VR Dataset}
  \Description{}
  \label{fig:ML}
\end{figure}

In this study, we demonstrate the applicability of the WARM-VR dataset through three use cases, for which we provide experimental results.
\begin{enumerate}
  \item \textbf{Investigating the Effect of Olfactory Stimuli:}
  In our previous work \cite{Valdivieso2025}, we demonstrated using the WARM-VR dataset that olfactory stimuli can subconsciously enhance relaxation in VR. To evaluate this effect, we analyzed the HRV derived from the ECG signals collected from the Polar H10. Particular attention was given to the HF band (0.15–0.4 Hz), which reflects parasympathetic nervous system activity linked to relaxation \cite{JacobRodrigues2023}. We then computed the Average HF values across all participants and compared them between experimental phases: the stress test and relaxation sessions with and without olfactory stimulation.
  \item \textbf{Affect Measurement via Machine Learning :}
  To demonstrate the applicability of the WARM-VR dataset in machine learning tasks, we developed several deep learning models to classify Valence, Arousal, and Relaxation states into binary categories (high vs. low). We chose to work with BVP signals from the E4 wristband. Figure \ref{fig:ML} shows the workflow. The ground truth labels were generated from participants' self-reported SAM and RRS scores, as outlined in Section \ref{ground-truth}.
    \begin{enumerate}
        \item \textbf{Signal Preprocessing:} To reduce noise introduced during setup, the first and last 5 seconds of each recording were trimmed. Since only one stress session included labels and each participant experienced two relaxation sessions, we only used the VR session with olfactory stimulation to avoid overfitting, as including both relaxation sessions would disproportionately increase the number of relaxation samples and lead to class imbalance. Thus, each participant contributed data from three conditions: baseline, stress, and olfactory-enhanced relaxation.
        \\\textbf{Signal Filtering and Segmentation:} A third-order Butterworth bandpass filter (0.7–3.7 Hz) was applied to the raw BVP signals to isolate the heart rate frequency range (40–220 bpm) and suppress noise, following the method in \cite{Salehizadeh2015}. The filtered signals were then segmented using a 60-second sliding window with a 5-second overlap, consistent with \cite{Rashid2021}. Each segment was assigned the same binary relaxation label as the session it originated from, based on the participant’s self-reported responses for that condition.
        \\\textbf{Standardization:} To normalize input values across subjects and sessions, each segment was standardized using Z-score normalization before being fed into the neural network.
        \item \textbf{Label Preparation:} Valence, Arousal, and Relaxation labels were binarized for classification. Segments with scores greater than 5 were labeled as high (1), while those with scores of 5 or lower were labeled as low (0).
        \item \textbf{Model Training:}
        \\\textbf{CNN Model Architecture:} We employed a 1D Convolutional Neural Network (CNN) due to its demonstrated effectiveness in affect recognition from BVP signals, as shown in previous works such as Lee M. et al. \cite{Lee2020} \cite{Lee2019} and Rashid N. et al \cite{Rashid2021}. The model consists of three convolutional blocks. The first two layers follow the architecture introduced in \cite{AlghoulTCN}. Where the first layer uses 8 filters with a kernel size of 64 and \textit{stride} of 4, followed by \textit{ReLU} activation, batch normalization, max pooling, and a 30\% \textit{dropout}. The second layer applies 16 filters (kernel size 32, \textit{stride} 2) with the same post-processing. We added a third layer that uses 8 filters (kernel size 16, \textit{stride} 1), again followed by batch normalization, max pooling, and \textit{dropout}. A Softmax-activated dense layer performs the final binary classification into high or low relaxation. 
        \\\textbf{Hybrid Model Architectures:} We also implemented several model variants by combining the first two convolutional layers (as described above) with either an LSTM or GRU layer. We experimented with 12 and 32 units and with Bidirectional models. The results are discussed in Section \ref{results}.
        The CNN and the Hybrid models’ results serve as a benchmark for future work in affect state estimation from wearable physiological signals within immersive VR settings.
        \\\textbf{Transformer Architectures:} As Transformers have become the state-of-the-art in many fields, we made sure to include the transformer architecture in the benchmark. For the Transformer-based models, we adapted an encoder-style architecture to 1D PPG segments by first converting the signal into patch embeddings. A 1D convolutional patch-embedding layer (kernel size 32, \textit{stride} 32) produces non-overlapping tokens of dimension \textit{dmodel}, to which learned positional embeddings are added. We evaluated two configurations: a lightweight Transformer with \textit{dmodel} =32 and 4 attention \textit{head}, and a higher-capacity Transformer-v2 with \textit{dmodel} = 64 and 2 \textit{heads}. The embedded sequence is passed through a single Transformer encoder block consisting of multi-head self-attention, residual connections, layer normalization, and a feed-forward network with \textit{GELU} activation (expansion ratio 2). \textit{Dropout} (0.3) is applied to both the attention and feed-forward outputs. Global average pooling aggregates the token representations, and a final Softmax layer outputs the binary class prediction.
        \\\textbf{Training:} To evaluate performance, we used a subject-independent 5-fold cross-validation scheme, ensuring that data from any given participant appeared exclusively in either the training or test set. This approach mirrors real-world conditions in which models must handle users they have never encountered before. To be consistent, all the models were trained with a batch size of 512, a learning rate of 0.001, 350 epochs, and early stopping of 80. We addressed label imbalance using per-fold class weighting, derived from the training data, and applied to the loss. This increased the influence of minority classes during learning and improved sensitivity to infrequent affect states. This was applied instead of the Binary cross-entropy loss function.
        \item \textbf{Performance Metrics:} We employed a set of metrics that capture not only overall accuracy but also robustness to class imbalance, an essential factor in affect-related applications. While accuracy offers a broad measure of correctness, it overlooks the distribution of misclassifications, which can be critical when the cost of errors varies across classes. The unweighted F1 score, commonly used in imbalanced datasets, treats all classes equally and emphasizes performance on minority class predictions. In addition to the F1 score, the Area Under the Curve (AUC) is a well-established metric known to outperform accuracy in many contexts \cite{Ling2003} \cite{Bradley1994}\cite{Halimu2019} and has recently been applied to affect recognition \cite{AlghoulTCN}. 
        By combining AUC, F1 score, and accuracy, we offer a comprehensive evaluation of model performance in terms of generalization, robustness, and fairness across all classes.
    \end{enumerate}
  \item \textbf{Statistical Analysis of Subjective Responses:}
  To validate the study protocol, we evaluated the self-reported questionnaires (TABLE \ref{fig:3} and \ref{table:4}). This helped us verify that the experimental conditions effectively manipulated the subjects' affective states as intended.
\end{enumerate}

\section{Results}
\label{results}

\subsection{Investigating the Effects of Olfactory Stimuli}
As shown in our previous work \cite{Valdivieso2025}, the results of the HRV analysis, specifically the HF component, show a significant increase in values during relaxation sessions compared to stress-inducing arithmetic tests. Notably, relaxation with olfactory stimulation (beach scent) led to a 109\% increase in HF, whereas relaxation without scent showed a 44\% increase. This difference was statistically significant (p = 0.002), indicating that olfactory stimuli substantially enhanced physiological relaxation beyond the effects of VR alone.

\subsection{Affect Measurement via Machine Learning}
The results of the binary classification task for the different affect states are presented in TABLE \ref{table:2}.

\begin{enumerate}
  \item \textbf{Valence Classification:}
   For the valence task, the baseline CNN and CNN-Bi-GRU-12 achieved the strongest overall performance, each reaching the highest average F1-score (0.63) and the highest AUC (0.69). Although CNN-Bi-LSTM-12 reached the highest validation accuracy (0.70), its performance was limited by a very low F1-0 score, indicating poor detection of the low-valence class. As a result, LSTM-based variants produced the weakest overall balance between classes. Transformer models showed stable but moderate performance, with average F1-scores of 0.59 and AUC values of 0.62 and 0.63. Overall, these findings suggest that while some hybrid architectures achieve higher accuracy, the simpler CNN—and its GRU-based extension—provide more reliable and balanced performance for valence prediction, suggesting the potential predominance of local temporal patterns over longer-range dependencies in BVP-based valence estimation.
  \item \textbf{Arousal Classification:}
   Arousal remained more challenging to discriminate than valence, with average F1-scores generally lower across architectures. The Transformer with dmodel=32, 4 heads achieved the best balanced performance, obtaining the highest average F1-score (0.59) and the strongest F1-0 (0.54), indicating improved detection of the low-arousal class. The Transformer-v2 (dmodel=64, 2 heads) and the baseline CNN followed closely, both reaching an average F1-score of 0.58 and sharing the highest AUC (0.68). CNN baseline also achived the highest Validation accuracy of 0.70. LSTM and GRU based models achived the worst compared to CNN only and transformers models. These results suggest that attention-based models provide a modest advantage for arousal estimation by better compensating for the class imbalance between low- and high-arousal states.
   \item \textbf{Relaxation Classification:}
   In the relaxation task,the overall performance was higher than for arousal, with several models achieving strong and well-balanced metrics. The baseline CNN obtained the best validation accuracy (0.71) and the highest F1-1 (0.79), but its lower F1-0 (0.46) indicates a bias toward the high-relaxation class. In contrast, CNN-Bi-GRU-12 achieved the most balanced results, with the highest F1-0 (0.59), the best average F1-score (0.64), and the top AUC (0.69). The Transformer with dmodel =32 and 4 heads matched this average F1 (0.64) and reached a competitive AUC of 0.68, while also maintaining strong F1-1 (0.76). These findings suggest that, for relaxation estimation, GRU-based and attention-based architectures provide a modest advantage over the plain CNN by improving detection of the low-relaxation state while preserving high performance on the high-relaxation class.
\end{enumerate}

\begin{table*}
  \caption{AVERAGE of SUBJECT INDEPENDENT 5-FOLD CROSS-VALIDATION RESULTS FOR BINARY CLASSIFICATION AFFECT STATES (LOW VS. HIGH) USING BVP DATA ACROSS VARIOUS MODELS}
  \label{table:2}
  \begin{tabular}{lccccc}
    \toprule
    \textbf{Model} & \textbf{Test Acc.} & \textbf{F1-score 0} & \textbf{F1-score 1} & \textbf{Ave. F1-score} & \textbf{AUC}\\
    \midrule
    \multicolumn{6}{c}{\textbf{Valence}} \\
    \hline
    \textbf{CNN} & 0.69 & \textbf{0.50} & 0.76 & \textbf{0.63} & \textbf{0.69} \\
    \textbf{CNN-GRU(12)} & 0.64 & 0.44 & 0.73 & 0.59 & 0.63 \\
    \textbf{CNN-GRU(32)} & 0.64 & 0.46 & 0.71 & 0.59 & 0.62 \\
    \textbf{CNN-Bi-GRU(12)} & 0.68 & 0.49 & 0.76 & \textbf{0.63} & \textbf{0.69} \\
    \textbf{CNN-LSTM(12)} & 0.67 & 0.35 & 0.78 & 0.57 & 0.57 \\
    \textbf{CNN-LSTM(32)} & 0.67 & 0.36 & 0.75 & 0.56 & 0.57 \\
    \textbf{CNN-Bi-LSTM(12)} & \textbf{0.70} & 0.38 & \textbf{0.80} & 0.59 & 0.59 \\
    \textbf{Transformer(32,4)} & 0.68 & 0.40 & 0.77 & 0.59 & 0.62 \\
    \textbf{Transformer-v2(64,2)} & 0.68 & 0.40 & 0.78 & 0.59 & 0.63 \\
    \hline
    \multicolumn{6}{c}{\textbf{Arousal}} \\
    \hline
    \textbf{CNN} & \textbf{0.70} & 0.47 & 0.68 & 0.58 & \textbf{0.68} \\
    \textbf{CNN-GRU(12)} & 0.63 & 0.32 & 0.68 & 0.50 & 0.62 \\
    \textbf{CNN-GRU(32)} & 0.67 & 0.40 & \textbf{0.70} & 0.55 & 0.67 \\
    \textbf{CNN-Bi-GRU(12)} & 0.64 & 0.38 & 0.69 & 0.54 & 0.59 \\
    \textbf{CNN-LSTM(12)} & 0.61 & 0.48 & 0.55 & 0.52 & 0.57 \\
    \textbf{CNN-LSTM(32)} & 0.66 & 0.40 & 0.68 & 0.54 & 0.63 \\
    \textbf{CNN-Bi-LSTM(12)} & 0.66 & 0.40 & 0.68 & 0.54 & 0.63 \\
    \textbf{Transformer(32,4)} & 0.62 & \textbf{0.54} & 0.63 & \textbf{0.59} & 0.65 \\
    \textbf{Transformer-v2(64,2)} & 0.63 & 0.50 & 0.65 & 0.58 & \textbf{0.68} \\
    \hline
    \multicolumn{6}{c}{\textbf{Relaxation}} \\
    \hline
    \textbf{CNN} & \textbf{0.71} & 0.46 & \textbf{0.79} & 0.63 & 0.60 \\
    \textbf{CNN-GRU(12)} & 0.65 & 0.56 & 0.68 & 0.62 & 0.64 \\
    \textbf{CNN-GRU(32)} & \textbf{0.62} & 0.54 & 0.66 & 0.60 & 0.63 \\
    \textbf{CNN-Bi-GRU(12)} & 0.66 & \textbf{0.59} & 0.69 & \textbf{0.64} & \textbf{0.69} \\
    \textbf{CNN-LSTM(12)} & 0.62 & 0.48 & 0.69 & 0.59 & 0.61 \\
    \textbf{CNN-LSTM(32)} & 0.68 & 0.36 & 0.78 & 0.57 & 0.57 \\
    \textbf{CNN-Bi-LSTM(12)} & 0.68 & 0.48 & 0.75 & 0.62 & 0.63 \\
    \textbf{Transformer(32,4)} & 0.68 & 0.52 & 0.76 & \textbf{0.64} & 0.68 \\
    \textbf{Transformer-v2(64,2)} & 0.67 & 0.51 & 0.74 & 0.63 & 0.67 \\
    \hline
    \bottomrule
  \end{tabular}
\end{table*}

\subsection{Statistical Analysis of Subjective Responses}

\begin{enumerate}
  \item \textbf{SAM and RRS analysis:}
    As shown in Table \ref{table:3}, subjective questionnaire analysis confirmed that the stress induction phase was effective in elevating stress before the VR relaxation phases. relaxation’s mean scores from the RRS decreased (Group A: 6.80 to 4.33; Group B: 7.69 to 2.84) between the baseline and stress phases, then increased (Group A: 4.33 to 7.26; Group B: 2.84 to 7.30) during the VR relaxation phases. To evaluate the statistical significance of these changes, a one-way repeated measures ANOVA followed by post-hoc paired t-tests was conducted separately for Group A and Group B. The results across the three phases revealed significant effects on arousal (Group A: p=0.0017; Group B: p=0.0009) and relaxation (Group A: p=0.0038; Group B: p=0.00001) in both groups, and on valence in Group B (p=0.0098). Post-hoc paired t-tests showed that, consistent with our expectations, arousal significantly increased from Baseline to Stress induction phases (Group A: p=0.0008; Group B: p=0.0059) and decreased from Stress induction to VR relaxation phases (Group A: p=0.0045; Group B: p=0.0041) in both groups. Additionally, relaxation significantly decreased from Baseline to Stress induction phases (Group A: p=0.0283; Group B: p=0.0017) and increased from Stress induction to VR relaxation phases (Group A: p=0.00001; Group B: p=0.00005). For Group B, valence did not change significantly from Baseline to Stress induction phases, but it did from Stress induction to VR relaxation phases (p=0.0049). These results align with our expectations that during stress induction, we expected arousal to increase and both relaxation and valence to decrease; conversely, during VR sessions, we expected valence and relaxation to increase and arousal to decrease.
  \item \textbf{Post-experiment questions analysis:}
   Most of the participants (74.2\%) confirmed that they would use the VR sessions in our study as a relaxation technique. This high percentage of positive responses supports the feasibility of deploying fully immersive VR-based systems for mental well-being applications. TABLE \ref{table:4} also shows the Mean and SD values of question 2, where we asked the participants to rate how immersive the VR beach environment with olfactory stimuli was compared to one without. Participants rated the olfactory-enhanced session as more immersive, with a Mean score of 3.62 (SD = 0.70), approaching the “very immersive” score of 4. These results indicate that olfactory input contributes meaningfully to the sense of presence in VR, aligning to enhance realism and affective impact in immersive systems.
\end{enumerate}

\begin{table*}
  \caption{AVERAGE 5-FOLD CROSS-VALIDATION RESULTS FOR BINARY CLASSIFICATION AFFECT STATES (LOW VS. HIGH) USING BVP DATA ACROSS VARIOUS MODELS}
  \label{table:3}
  \begin{tabular}{lccc}
    \toprule
    \textbf{Group A} & \textbf{SAM - Valence} & \textbf{SAM - Arousal} & \textbf{RRS}\\
    \midrule
    Baseline phase & $6.13 \pm 2.06$ & $4.26 \pm 2.66$ & $6.80 \pm 2.56$ \\
    Stress induction phase & $5.66 \pm 2.69$ & $7.13 \pm 1.85$ & $4.33 \pm 3.02$ \\
    VR Relaxation phase & $7.13 \pm 1.58$ & $4.40 \pm 2.72$ & $7.26 \pm 1.80$ \\
    \hline
    \textbf{Group B} & \textbf{SAM - Valence} & \textbf{SAM - Arousal} & \textbf{RRS} \\
    \hline
    Baseline phase & $6.61 \pm 1.77$ & $4.76 \pm 2.83$ & $7.69 \pm 1.13$ \\
    Stress induction phase & $5.15 \pm 2.71$ & $7.76 \pm 1.12$ & $2.84 \pm 2.03$ \\
    VR Relaxation with Olfactory phase & $7.69 \pm 1.13$ & $4.23 \pm 3.14$ & $7.30 \pm 2.46$\\
    \bottomrule
  \end{tabular}
\end{table*}

\section{Conclusion}

In this paper, we introduced WARM-VR (Wearable Affect Recognition from Multisensory stimuli in Virtual Reality), a novel dataset designed to support research in multimodal affective computing through the integration of physiological signals, immersive virtual environments, and multisensory stimuli. Unlike existing datasets, WARM-VR was developed specifically to explore the effect of olfactory-enhanced VR experiences on affective states, particularly relaxation and stress. The dataset includes synchronized physiological measurement recordings from chest and wristband wearable sensors, alongside validated self-report measures of affective states.

We demonstrated two use cases for the dataset: (1) investigating the effect of olfactory stimuli on physiological indicators of relaxation using HRV analysis, and (2) Training a deep learning model to estimate binary Valence, Arousal and Relaxation levels from BVP data of the wristband. These examples illustrate the flexibility and potential of WARM-VR to support both traditional statistical analysis and machine learning approaches in affect recognition. 

WARM-VR fills a critical gap in existing datasets by providing multimodal data collected in a fully immersive and wearable setting. We believe this dataset will enable new research directions in affect-aware systems, stress monitoring, personalized well-being, and real-time biofeedback in virtual environments.

Future work will focus on expanding the dataset with additional participants, longer-term exposure scenarios, and a broader range of affective states to further support generalizable and real-world applications of affect recognition.

\begin{table*}
  \caption{POST-EXPERIMENT QUESTIONNAIRE ANSWERS}
  \label{table:4}
  \begin{tabular}{p{7cm}|p{5cm}}
    \toprule
    \textbf{Question} & \textbf{Answer}\\
    \midrule
    \textbf{Q1:} Will you use this as a relaxation technique? & Yes (73.4\%), No (13.3\%), and Maybe (13.3\%) \\
    \hline
    \textbf{Q2:} How immersive was the beach environment with scent compared to without scent (Use Likert scale from 1 to 5, with 1 being not at all immersive and 5 being the most immersive) & Mean and SD: 3.63$\pm$0.70 \\
    \bottomrule
  \end{tabular}
\end{table*}

\bibliographystyle{ACM-Reference-Format}
\bibliography{warm-ref}

@article{Lee2020,
   author = {Min Seop Lee and Yun Kyu Lee and Myo Taeg Lim and Tae Koo Kang},
   doi = {10.3390/app10103501},
   issn = {20763417},
   issue = {10},
   journal = {Applied Sciences (Switzerland)},
   month = {5},
   publisher = {MDPI AG},
   title = {Emotion recognition using convolutional neural network with selected statistical photoplethysmogram features},
   volume = {10},
   year = {2020}
}

@inproceedings{Rashid2021,
   author = {Nafiul Rashid and Luke Chen and Manik Dautta and Abel Jimenez and Peter Tseng and Mohammad Abdullah Al Faruque},
   doi = {10.1109/EMBC46164.2021.9630576},
   isbn = {9781728111797},
   issn = {1557170X},
   booktitle = {Proceedings of the Annual International Conference of the IEEE Engineering in Medicine and Biology Society, EMBS},
   pages = {2374-2377},
   pmid = {34891759},
   publisher = {Institute of Electrical and Electronics Engineers Inc.},
   title = {Feature Augmented Hybrid CNN for Stress Recognition Using Wrist-based Photoplethysmography Sensor},
   year = {2021}
}

@article{Lee2019,
   author = {Min Seop Lee and Yun Kyu Lee and Dong Sung Pae and Myo Taeg Lim and Dong Won Kim and Tae Koo Kang},
   doi = {10.3390/app9163355},
   issn = {20763417},
   issue = {16},
   journal = {Applied Sciences (Switzerland)},
   month = {8},
   publisher = {MDPI AG},
   title = {Fast emotion recognition based on single pulse PPG signal with convolutional neural network},
   volume = {9},
   year = {2019}
}

@article{Salehizadeh2015,
   author = {Seyed Salehizadeh and Duy Dao and Jeffrey Bolkhovsky and Chae Cho and Yitzhak Mendelson and Ki Chon},
   doi = {10.3390/s16010010},
   issn = {1424-8220},
   issue = {1},
   journal = {Sensors},
   month = {12},
   pages = {10},
   title = {A Novel Time-Varying Spectral Filtering Algorithm for Reconstruction of Motion Artifact Corrupted Heart Rate Signals During Intense Physical Activities Using a Wearable Photoplethysmogram Sensor},
   volume = {16},
   year = {2015}
}

@article{CamaraD,
   author = {Dayne R Camara and Richard E Hicks},
   doi = {10.5176/2345-7929_4.2.100},
   title = {USING VIRTUAL REALITY TO REDUCE STATE ANXIETY AND STRESS IN UNIVERSITY STUDENTS: AN EXPERIMENT}
}

@article{Grenier2015,
   author = {Sébastien Grenier and Hélène Forget and Stéphane Bouchard and Sébastien Isere and Sylvie Belleville and Olivier Potvin and Marie-Ève Rioux and Mélissa Talbot},
   doi = {10.1017/S1041610214002300},
   issn = {1041-6102},
   issue = {7},
   journal = {International Psychogeriatrics},
   month = {7},
   pages = {1217-1225},
   title = {Using virtual reality to improve the efficacy of cognitive-behavioral therapy (CBT) in the treatment of late-life anxiety: preliminary recommendations for future research},
   volume = {27},
   year = {2015}
}

@article{Bradley1994,
   author = {Margaret M. Bradley and Peter J. Lang},
   doi = {10.1016/0005-7916(94)90063-9},
   issn = {00057916},
   issue = {1},
   journal = {Journal of Behavior Therapy and Experimental Psychiatry},
   month = {3},
   pages = {49-59},
   title = {Measuring emotion: The self-assessment manikin and the semantic differential},
   volume = {25},
   year = {1994}
}

@article{SAKAKIBARA1994,
   author = {MASAHITO SAKAKIBARA and SATOSHI TAKEUCHI and JUNICHIRO HAYANO},
   doi = {10.1111/j.1469-8986.1994.tb02210.x},
   issn = {0048-5772},
   issue = {3},
   journal = {Psychophysiology},
   month = {5},
   pages = {223-228},
   title = {Effect of relaxation training on cardiac parasympathetic tone},
   volume = {31},
   year = {1994}
}

@article{Ahmad2022,
   author = {Zeeshan Ahmad and Naimul Khan},
   doi = {10.3390/bioengineering9110688},
   issn = {2306-5354},
   issue = {11},
   journal = {Bioengineering},
   month = {11},
   pages = {688},
   title = {A Survey on Physiological Signal-Based Emotion Recognition},
   volume = {9},
   year = {2022}
}

@inbook{Ling2003,
   author = {Charles X. Ling and Jin Huang and Harry Zhang},
   doi = {10.1007/3-540-44886-1_25},
   pages = {329-341},
   title = {AUC: A Better Measure than Accuracy in Comparing Learning Algorithms},
   year = {2003}
}

@article{AlOsman2016,
   author = {Hussein Al Osman and Haiwei Dong and Abdulmotaleb El Saddik},
   doi = {10.1109/ACCESS.2016.2548980},
   issn = {21693536},
   journal = {IEEE Access},
   pages = {1274-1286},
   publisher = {Institute of Electrical and Electronics Engineers Inc.},
   title = {Ubiquitous Biofeedback Serious Game for Stress Management},
   volume = {4},
   year = {2016}
}

@article{Dademir2023,
   author = {Yaşar Daşdemir},
   doi = {10.3390/diagnostics13223437},
   issn = {20754418},
   issue = {22},
   journal = {Diagnostics},
   month = {11},
   publisher = {Multidisciplinary Digital Publishing Institute (MDPI)},
   title = {Classification of Emotional and Immersive Outcomes in the Context of Virtual Reality Scene Interactions},
   volume = {13},
   year = {2023}
}

@inproceedings{Schmidt2018,
   author = {Philip Schmidt and Attila Reiss and Robert Duerichen and Kristof Van Laerhoven},
   doi = {10.1145/3242969.3242985},
   isbn = {9781450356923},
   booktitle = {ICMI 2018 - Proceedings of the 2018 International Conference on Multimodal Interaction},
   month = {10},
   pages = {400-408},
   publisher = {Association for Computing Machinery, Inc},
   title = {Introducing WeSAD, a multimodal dataset for wearable stress and affect detection},
   year = {2018}
}

@article{Jin2022,
   author = {Ye Ji Jin and Erkinov Habibilloh and Ye Seul Jang and Taejun An and Donghyun Jo and Saron Park and Won Du Chang},
   doi = {10.3390/app12136544},
   issn = {20763417},
   issue = {13},
   journal = {Applied Sciences (Switzerland)},
   month = {7},
   publisher = {MDPI},
   title = {A Photoplethysmogram Dataset for Emotional Analysis},
   volume = {12},
   year = {2022}
}

@article{Koelstra2012,
   author = {Sander Koelstra and Christian Mühl and Mohammad Soleymani and Jong Seok Lee and Ashkan Yazdani and Touradj Ebrahimi and Thierry Pun and Anton Nijholt and Ioannis Patras},
   doi = {10.1109/T-AFFC.2011.15},
   issn = {19493045},
   issue = {1},
   journal = {IEEE Transactions on Affective Computing},
   month = {1},
   pages = {18-31},
   title = {DEAP: A database for emotion analysis; Using physiological signals},
   volume = {3},
   year = {2012}
}

@article{Dademir2022,
   author = {Yaşar Daşdemir},
   doi = {10.1016/j.bspc.2022.103942},
   issn = {17468108},
   journal = {Biomedical Signal Processing and Control},
   month = {9},
   publisher = {Elsevier Ltd},
   title = {Cognitive investigation on the effect of augmented reality-based reading on emotion classification performance: A new dataset},
   volume = {78},
   year = {2022}
}

@article{Xue2022,
   author = {Jingyi Xue and Jinqin Wang and Shiang Hu and Ning Bi and Zhao Lv},
   doi = {10.1109/TIM.2022.3149116},
   issn = {15579662},
   journal = {IEEE Transactions on Instrumentation and Measurement},
   publisher = {Institute of Electrical and Electronics Engineers Inc.},
   title = {OVPD: Odor-Video Elicited Physiological Signal Database for Emotion Recognition},
   volume = {71},
   year = {2022}
}

@article{Raheel2021,
   author = {Aasim Raheel and Muhammad Majid and Syed Muhammad Anwar},
   doi = {10.1016/j.inffus.2020.08.007},
   issn = {15662535},
   journal = {Information Fusion},
   month = {1},
   pages = {37-49},
   publisher = {Elsevier B.V.},
   title = {DEAR-MULSEMEDIA: Dataset for emotion analysis and recognition in response to multiple sensorial media},
   volume = {65},
   year = {2021}
}

@article{Igarashi2014,
   author = {Miho Igarashi and Harumi Ikei and Chorong Song and Yoshifumi Miyazaki},
   doi = {10.1016/j.ctim.2014.09.003},
   issn = {09652299},
   issue = {6},
   journal = {Complementary Therapies in Medicine},
   month = {12},
   pages = {1027-1031},
   title = {Effects of olfactory stimulation with rose and orange oil on prefrontal cortex activity},
   volume = {22},
   year = {2014}
}

@article{Benbow2019,
   author = {Amanda A. Benbow and Page L. Anderson},
   doi = {10.1016/j.janxdis.2018.06.006},
   issn = {08876185},
   journal = {Journal of Anxiety Disorders},
   month = {1},
   pages = {18-26},
   title = {A meta-analytic examination of attrition in virtual reality exposure therapy for anxiety disorders},
   volume = {61},
   year = {2019}
}

@misc{polar,
  title        = {Polar H10 Heart Rate Sensor},
  howpublished = {\url{https://www.polar.com/ca-en/sensors/h10-heart-rate-sensor}},
  year         = {2025},
  note         = {Accessed: 2025-02-26}
}

@misc{e4,
  title        = {Empatica E4 Wristband for Research},
  howpublished = {\url{https://www.empatica.com/research/e4/}},
  year         = {2025},
  note         = {Accessed: 2025-02-25}
}

@misc{metaQ,
  title        = {Meta Quest 2 - Virtual Reality Headset},
  howpublished = {\url{https://www.meta.com/ca/quest/}},
  year         = {2025},
  note         = {Accessed: 2025-01-19}
}

@article{Li2022,
   author = {Dahua Li and Zhiyi Yang and Fazheng Hou and Qiaoju Kang and Shuang Liu and Yu Song and Qiang Gao and Enzeng Dong},
   doi = {10.1109/TIM.2022.3147882},
   issn = {0018-9456},
   journal = {IEEE Transactions on Instrumentation and Measurement},
   pages = {1-11},
   title = {EEG-Based Emotion Recognition With Haptic Vibration by a Feature Fusion Method},
   volume = {71},
   year = {2022}
}

@article{Milstein2020,
   author = {Nir Milstein and Ilanit Gordon},
   doi = {10.3389/fnbeh.2020.00148},
   issn = {1662-5153},
   journal = {Frontiers in Behavioral Neuroscience},
   month = {8},
   title = {Validating Measures of Electrodermal Activity and Heart Rate Variability Derived From the Empatica E4 Utilized in Research Settings That Involve Interactive Dyadic States},
   volume = {14},
   year = {2020}
}

@techReport{kutt,
   author = {Krzysztof Kutt and Pawel Wegrzyn and Szymon Bobek and Grzegorz J Nalepa and Jan Argasi«ski and Paweª W¦grzyn and Mateusz},
   title = {Affective Computing Experiments in Virtual Reality with Wearable Sensors. Methodological considerations and preliminary results Aective Computing Experiments in Virtual Reality with Wearable Sensors. Methodological considerations and preliminary results},
   url = {https://www.researchgate.net/publication/313841760}
}

@article{Birckhead2019,
   author = {Brandon Birckhead and Carine Khalil and Xiaoyu Liu and Samuel Conovitz and Albert Rizzo and Itai Danovitch and Kim Bullock and Brennan Spiegel},
   doi = {10.2196/11973},
   issn = {2368-7959},
   issue = {1},
   journal = {JMIR Mental Health},
   month = {1},
   pages = {e11973},
   title = {Recommendations for Methodology of Virtual Reality Clinical Trials in Health Care by an International Working Group: Iterative Study},
   volume = {6},
   year = {2019}
}

@inproceedings{Valdivieso2025,
   author = {Yasmin Elsaddik Valdivieso and Mohd Faisal and Karim Alghoul and Monireh Monica Vahdati and Kamran Gholizadeh Hamlabadi and Fedwa Laamarti and Hussein Al Osman and Abdulmotaleb El Saddik},
   doi = {10.1109/MeMeA65319.2025.11068102},
   isbn = {979-8-3315-2347-3},
   booktitle = {2025 IEEE Medical Measurements \&amp; Applications (MeMeA)},
   month = {5},
   pages = {1-6},
   publisher = {IEEE},
   title = {The Potential of Olfactory Stimuli in Stress Reduction Through Virtual Reality},
   year = {2025}
}

@article{Lv2024,
   author = {Zhihan Lv and Fabio Poiesi and Qi Dong and Jaime Lloret and Houbing Song},
   doi = {10.1145/3605151},
   issn = {1551-6857},
   issue = {2},
   journal = {ACM Transactions on Multimedia Computing, Communications, and Applications},
   month = {2},
   pages = {1-5},
   title = {Special Issue on Deep Learning for Intelligent Human Computer Interaction},
   volume = {20},
   year = {2024}
}

@article{Zhao2019,
   author = {Sicheng Zhao and Shangfei Wang and Mohammad Soleymani and Dhiraj Joshi and Qiang Ji},
   doi = {10.1145/3363560},
   issn = {1551-6857},
   issue = {3s},
   journal = {ACM Transactions on Multimedia Computing, Communications, and Applications},
   month = {11},
   pages = {1-32},
   title = {Affective Computing for Large-scale Heterogeneous Multimedia Data},
   volume = {15},
   year = {2019}
}

@article{Zhang2025,
   author = {Xinjie Zhang and Tenggan Zhang and Lei Sun and Jinming Zhao and Qin Jin},
   doi = {10.1145/3723005},
   issn = {1551-6857},
   issue = {7},
   journal = {ACM Transactions on Multimedia Computing, Communications, and Applications},
   month = {7},
   pages = {1-28},
   title = {Exploring Interpretability in Deep Learning for Affective Computing: A Comprehensive Review},
   volume = {21},
   year = {2025}
}

@article{Yin2022,
   author = {Guanghao Yin and Shouqian Sun and Dian Yu and Dejian Li and Kejun Zhang},
   doi = {10.1145/3490686},
   issn = {1551-6857},
   issue = {3},
   journal = {ACM Transactions on Multimedia Computing, Communications, and Applications},
   month = {8},
   pages = {1-23},
   title = {A Multimodal Framework for Large-Scale Emotion Recognition by Fusing Music and Electrodermal Activity Signals},
   volume = {18},
   year = {2022}
}

@article{Chen2022,
   author = {Min Chen and Wenjing Xiao and Miao Li and Yixue Hao and Long Hu and Guangming Tao},
   doi = {10.1145/3462763},
   issn = {1551-6857},
   issue = {1s},
   journal = {ACM Transactions on Multimedia Computing, Communications, and Applications},
   month = {2},
   pages = {1-18},
   title = {A Multi-feature and Time-aware-based Stress Evaluation Mechanism for Mental Status Adjustment},
   volume = {18},
   year = {2022}
}

@article{Vahdati2025,
   author = {Monireh (Monica) Vahdati and Fedwa Laamarti and Abdulmotaleb El Saddik},
   doi = {10.1145/3705320},
   issn = {1551-6857},
   issue = {7},
   journal = {ACM Transactions on Multimedia Computing, Communications, and Applications},
   month = {7},
   pages = {1-36},
   title = {Meta-Review of Wearable Devices for Healthcare in the Metaverse},
   volume = {21},
   year = {2025}
}

@article{Rashed2025,
   author = {Ammar Rashed and Shervin Shirmohammadi and Ihab Amer and Mohamed Hefeeda},
   doi = {10.1145/3722116},
   issn = {1551-6857},
   issue = {7},
   journal = {ACM Transactions on Multimedia Computing, Communications, and Applications},
   month = {7},
   pages = {1-33},
   title = {A Review of Player Engagement Estimation in Video Games: Challenges and Opportunities},
   volume = {21},
   year = {2025}
}

@article{Zhu2019,
   author = {Junjie Zhu and Yuxuan Wei and Yifan Feng and Xibin Zhao and Yue Gao},
   doi = {10.1145/3332374},
   issn = {1551-6857},
   issue = {3s},
   journal = {ACM Transactions on Multimedia Computing, Communications, and Applications},
   month = {11},
   pages = {1-18},
   title = {Physiological Signals-based Emotion Recognition via High-order Correlation Learning},
   volume = {15},
   year = {2019}
}

@article{Zhao2019v2,
   author = {Sicheng Zhao and Amir Gholaminejad and Guiguang Ding and Yue Gao and Jungong Han and Kurt Keutzer},
   doi = {10.1145/3233184},
   issn = {1551-6857},
   issue = {1s},
   journal = {ACM Transactions on Multimedia Computing, Communications, and Applications},
   month = {1},
   pages = {1-18},
   title = {Personalized Emotion Recognition by Personality-Aware High-Order Learning of Physiological Signals},
   volume = {15},
   year = {2019}
}

@article{Latifzadeh2024,
   author = {Kayhan Latifzadeh and Nima Gozalpour and V. Javier Traver and Tuukka Ruotsalo and Aleksandra Kawala-Sterniuk and Luis A Leiva},
   doi = {10.1145/3663669},
   issn = {1551-6857},
   issue = {10},
   journal = {ACM Transactions on Multimedia Computing, Communications, and Applications},
   month = {10},
   pages = {1-24},
   title = {Efficient Decoding of Affective States from Video-elicited EEG Signals: An Empirical Investigation},
   volume = {20},
   year = {2024}
}

@article{JacobRodrigues2023,
   author = {Mariana C. Jacob Rodrigues and Octavian Postolache and Francisco Cercas},
   doi = {10.1109/TIM.2023.3279881},
   issn = {0018-9456},
   journal = {IEEE Transactions on Instrumentation and Measurement},
   pages = {1-19},
   title = {The Influence of Stress Noise and Music Stimulation on the Autonomous Nervous System},
   volume = {72},
   year = {2023}
}

@inproceedings{Halimu2019,
   author = {Chongomweru Halimu and Asem Kasem and S. H. Shah Newaz},
   city = {New York, NY, USA},
   doi = {10.1145/3310986.3311023},
   isbn = {9781450366120},
   booktitle = {Proceedings of the 3rd International Conference on Machine Learning and Soft Computing},
   month = {1},
   pages = {1-6},
   publisher = {ACM},
   title = {Empirical Comparison of Area under ROC curve (AUC) and Mathew Correlation Coefficient (MCC) for Evaluating Machine Learning Algorithms on Imbalanced Datasets for Binary Classification},
   year = {2019}
}

@INPROCEEDINGS{AlghoulTCN,
  author={Alghoul, Karim and Al Osman, Hussein and El Saddik, Abdulmotaleb},
  booktitle={2025 IEEE International Instrumentation and Measurement Technology Conference (I2MTC)}, 
  title={Enhancing Generalization in PPG-Based Emotion Measurement with a CNN-TCN-LSTM Model}, 
  year={2025},
  volume={},
  number={},
  pages={1-6},
  doi={10.1109/I2MTC62753.2025.11079085}}

\end{document}